\documentclass[sigconf]{acmart}

\AtBeginDocument{%
  }

\copyrightyear{2025}
\acmYear{2025}
\setcopyright{cc}
\setcctype{by}
\acmConference[LAVA '25]{Proceedings of the 2nd International Workshop on Large Vision - Language Model Learning and Applications}{October 27--28, 2025}{Dublin, Ireland}
\acmBooktitle{Proceedings of the 2nd International Workshop on Large Vision - Language Model Learning and Applications (LAVA '25), October 27--28, 2025, Dublin, Ireland}\acmDOI{10.1145/3728483.3760194}
\acmISBN{979-8-4007-1840-3/2025/10}

\usepackage{subcaption}
\usepackage{multirow}
\begin{document}

\title{Enhancing Scientific Visual Question Answering via Vision-Caption aware Supervised Fine-Tuning}

\author{Janak Kapuriya}
\affiliation{%
\institution{Indraprastha Institute of Information Technology, Delhi}
\city{New Delhi}
\country{India}
}
\email{kapuriya22032@iiitd.ac.in}

\author{Anwar Shaikh}
\affiliation{%
\institution{Indraprastha Institute of Information Technology, Delhi}
\city{New Delhi}
\country{India}
}
\email{anwars@iiitd.ac.in}

\author{Arnav Goel}
\affiliation{%
\institution{Indraprastha Institute of Information Technology, Delhi}
\city{New Delhi}
\country{India}
}
\email{arnav21519@iiitd.ac.in}

\author{Medha Hira}
\affiliation{%
\institution{Indraprastha Institute of Information Technology, Delhi}
\city{New Delhi}
\country{India}
}
\email{medha21265@iiitd.ac.in}

\author{Apoorv Singh}
\affiliation{%
\institution{Indraprastha Institute of Information Technology, Delhi}
\city{New Delhi}
\country{India}
}
\email{apoorv17027@iiitd.ac.in}

\author{Jay Saraf}
\affiliation{%
\institution{Indraprastha Institute of Information Technology, Delhi}
\city{New Delhi}
\country{India}
}
\email{jay20438@iiitd.ac.in}

\author{Sanjana}
\affiliation{%
\institution{Indraprastha Institute of Information Technology, Delhi}
\city{New Delhi}
\country{India}
}
\email{sanjana21094@iiitd.ac.in}

\author{Vaibhav Nauriyal}
\affiliation{%
\institution{Indraprastha Institute of Information Technology, Delhi}
\city{New Delhi}
\country{India}
}
\email{vaibhav22129@iiitd.ac.in}

\author{Avinash Anand}
\affiliation{%
\institution{Indraprastha Institute of Information Technology, Delhi}
\city{New Delhi}
\country{India}
}
\email{avinasha@iiitd.ac.in}

\author{Zhengkui Wang}
\affiliation{%
\institution{Singapore Institute of Technology}
\city{Singapore}
\country{Singapore}
}

\email{zhengkui.wang@singaporetech.edu.sg}

\author{Rajiv Ratn Shah}
\affiliation{%
\institution{Indraprastha Institute of Information Technology, Delhi}
\city{New Delhi}
\country{India}
}
\email{rajivratn@iiitd.ac.in}

\renewcommand{\shortauthors}{Janak Kapuriya et al.}

\begin{abstract}
 In this study, we introduce Vision-Caption aware Supervised Fine-Tuning (VCASFT), a novel learning paradigm designed to enhance the performance of smaller Vision Language Models(VLMs) on scientific visual question answering(VQA) tasks. VCASFT leverages image captions as zero-shot prompts alongside question-answer pairs and instruction-tunes models to yield significant performance improvements. To comprehensively evaluate VCASFT, we benchmark it on ScienceQA, which consists of questions across diverse languages, subjects, and fields, demonstrating its adaptability and effectiveness in a variety of educational contexts. Additionally, to further demonstrate the effectiveness of this technique on low-resource languages, we developed HiSciVQA, a dataset comprising 2,245 high-quality, hand-annotated Hindi multimodal Q\&A pairs. This dataset addresses the critical need for low-resource language Q\&A datasets and serves as a foundation for testing VCASFT. Additionally, we introduce a novel LLM-based evaluation scheme to evaluate VLMs on HiSciVQA which offers deeper insights into model effectiveness surpassing traditional n-gram matching accuracy metrics. We are committed to advancing the field by open-sourcing all code files and the HiSciVQA dataset for the research community.
\end{abstract}



\maketitle

\begin{figure*}
    \centering
    \includegraphics[width=0.65\textwidth]{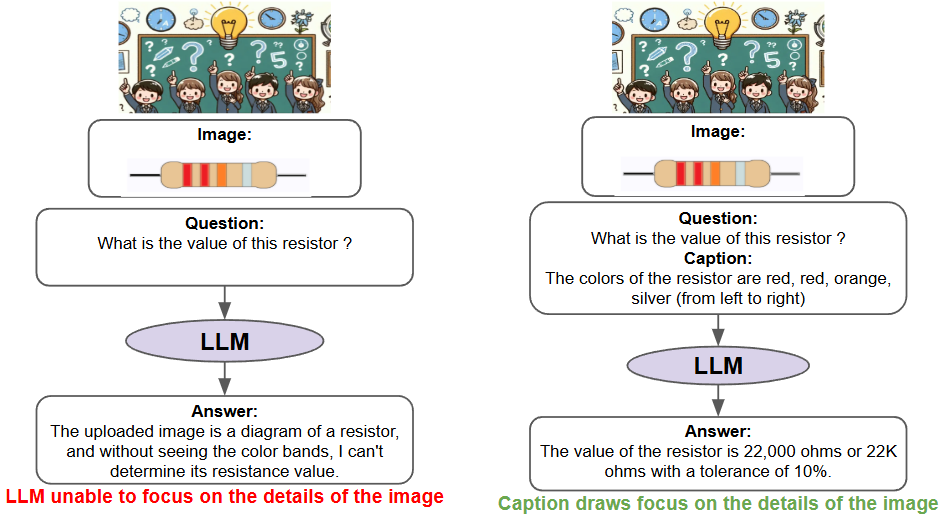}
    \caption{Outputs of GPT-4V w/o and with Image Caption respectively.}
    \label{fig:explanation}
\end{figure*}

\section{Introduction}
Transformer-based \citep{vaswani2017attention} Vision Language Models (VLMs) have emerged in recent times due to their superior capabilities in zero-shot generalization, image-based question answering, text generation, and instruction-following \citep{ghosh2024exploring, zhang2024vision, yin2023survey, wu2023multimodal}. These models integrate visual image encoders with state-of-the-art LLMs such as GPT-4 \citep{achiam2023gpt}, Mistral \citep{agrawal2024pixtral}, and LLAMA \citep{dubey2024llama}, which demonstrate exceptional proficiency in mimicking human-like text generation. 

Inspite of these advancements, scientific question-answering remains a domain where novice auto regressive text generation has yielded fewer gains. Studies focused on enhancing language models on these tasks majorly focused on Math Word Problems (MWP) by employing reasoning, exploring symbolic representations, or utilizing graph-based models of questions \citep{zhou2022least, gaur2023reasoning, opedal2023world, he2023solving, kojima2022large}. Visual scientific word problems (VSWPs) are however different from MWPs as they demand an accurate grounding in a fact and emphasis on the information detailed in the adjoined image. Thus, the techniques for MWPs do not directly apply to SWPs \citep{10.1007/978-3-031-49601-1_4}. Since LLMs generate answers based on their extensive training, their vast parameters can sometimes falter in specialized areas, necessitating support through additional data and instruction tuning \citep{jiang2024instructiontunedlanguagemodelsbetter}. This prompts us in the direction to augment vision language models using domain-specific tuning.

Current VLMs often inadequately extract visual information for reasoning tasks, relying heavily on text inputs and exhibiting "blindness" to visual cues \citep{wang2024exploring, zhang2024mathverse, rahmanzadehgervi2024visionlanguagemodelsblind}. This limitation is prominent in smaller models with weaker vision encoders and in low-resource settings \citep{wang2024exploring}. We address this by generating descriptive captions for images (see Figure \ref{fig:explanation}) to enhance VLM adaptation for the scientific visual question answering (VQA) task, as detailed in Section \ref{sec:methodology}. Our method improves performance by approximately \textbf{8 percentage points} on smaller, 7-billion parameter models when tested on the ScienceQA dataset (Section \ref{sec:eng_results}). Additionally, we introduce a Hindi-language Science VQA (HiSciVQA) dataset in Section \ref{sec:dataset} to support VCASFT training of multilingual VLMs such as the Palo family, notably improving the smallest 1.7 billion parameter model's performance by nearly \textbf{15 points} (Section \ref{sec:hindi_results}).

Thus, the contributions of this research paper are \textbf{threefold} and they are described as follows:
\textbf{First:} We introduce the \textbf{Vision-Caption aware Supervised Fine-Tuning Technique (VCASFT)}, a novel instructional-tuning method designed to enhance the adaptability of smaller Vision Language Models (VLMs) for scientific VQA tasks by tuning the models with cues to better attend to the vision modality.
\textbf{Second:} We introduce \textbf{HiSciVQA}, the \textit{first Hindi language multimodal dataset} focused on high-school science, featuring a rich variety of question types and high-quality explanations, marking the first endeavor of its kind within this domain and will make this dataset publicly available for the research community.
\textbf{Third:} We introduce a new evaluation schema to enhance traditional accuracy assessments, offering deeper insights into model effectiveness for scientific Q\&A challenges. By benchmarking recently developed VLMs within our VCASFT framework, we lay the ground for building better reasoners upon these models.

\begin{figure*}[h]
  \centering
  \includegraphics[width=0.7\textwidth]{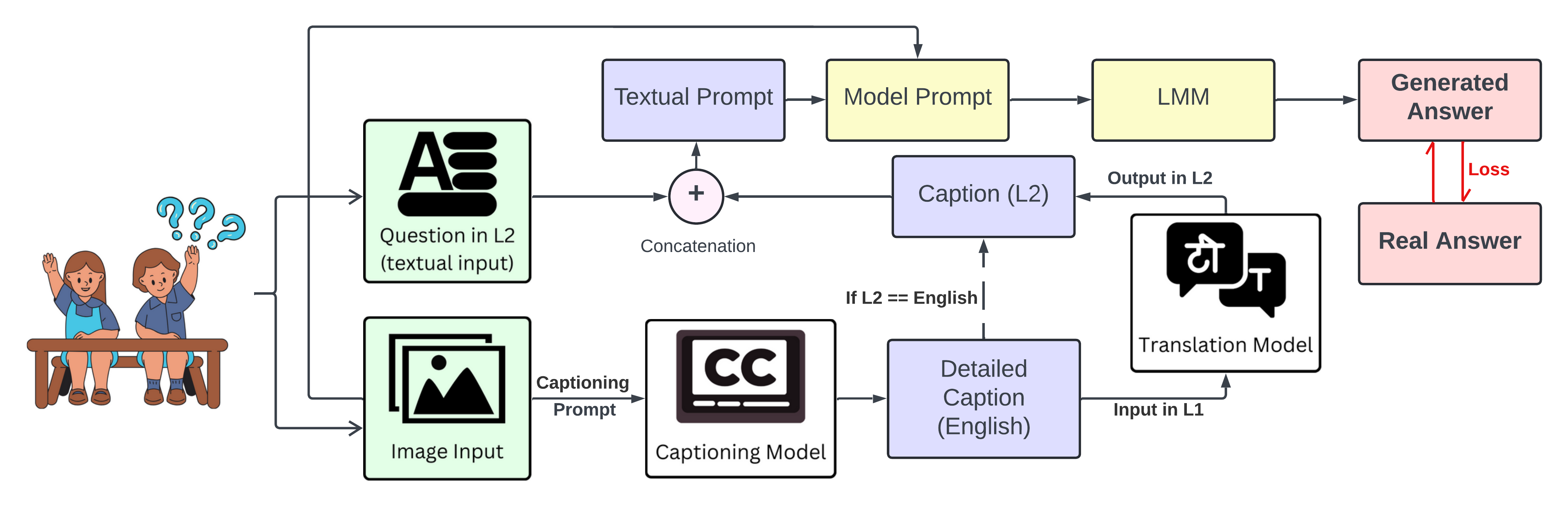}
  \caption{Pipeline showing our proposed Vision-Caption aware Supervised Fine-Tuning Framework (VCASFT).}
  \label{fig:VCASFT}
\end{figure*}

\section{Related Work} \label{related-work}
In our dataset exploration, we found several Mathematics and Science Q\&A datasets but observed a lack of high-quality, challenging scientific visual question-answering datasets. In Section 2.1 we discuss about existing scientific VQA datasets and Section 2.2 describes various vision-language models available for VQA.

\subsection{Existing Scientific VQA Datasets} 
Text-based datasets such as SciPhyRAG \citep{10.1007/978-3-031-49601-1_4} and those derived from NCERT exemplars \citep{10.1007/978-3-031-49601-1_5} provide extensive resources for high school science, featuring thousands of questions with step-by-step explanations and employing information retrieval methodologies. In contrast, open-domain science Q\&A datasets like ScienceQA \citep{lu2022learn} and SciQ \citep{welbl2017crowdsourcing} focus on basic fact-based multiple-choice questions. Computational challenges are represented in datasets such as SCIMAT \citep{kollepara2021scimat} and JEEBench \citep{arora2023have}, which are based on higher education syllabi but are limited by the lack of detailed explanations and scope, respectively. These limitations highlight the need for more comprehensive datasets that can enhance LLMs for scientific Q\&A. Furthermore, there is a critical shortage of datasets in low-resource Indic languages like Hindi, which is widely spoken, highlighting the need for resources that cater to diverse linguistic populations, especially in educational and computational Q\&A contexts.

\subsection{Vision-Language Models for VQA}
Initial approaches to VQA adopted data-driven strategies aimed at mapping questions and answers \citep{malinowski2014multi, agrawal2016vqa, ren2015exploring}. Whitehead et al. \citep{whitehead2021separating} enhanced VQA performance by improving generalization through contrastive learning, separating skills and concepts. Recent transformer-based models Vision Language Models (VLMs) using pre-trained image encoders and LLMs such as GPT-4 \citep{achiam2023gpt}, Gemini \citep{team2023gemini}, LLaVA \citep{liu2023llava}, Pixtral \citep{agrawal2024pixtral}, InternVL \citep{chen2024internvl} and Qwen2-VL \citep{yang2024qwen2} have advanced the integration of language and vision-based modalities improving the interaction between the two. PALO \citep{maaz2024palo} further enhances VLMs by offering visual language modelling in ten languages, including Hindi. 

Recent works focusing on adapting these models to scientific VQA try to enhance their reasoning ability through enhanced prompting and instructing-tuning techniques \citep{lan2023improving, yan2024errorradar}. The \textit{blindness} of these models is acknowledged with existing work using captions as prompts for open-domain VQA \citep{Ozdemir_2024_CVPR} and enhancing the image encoder for better diagram encoding \citep{zhang2024mavis}. Additional work focuses on improved data-driven pre-training and alignment endeavors to induce visual comprehension and reasoning \citep{peng2024multimath, jia2024describe}. However, we observe a gap in this study of using image descriptions in the alignment process to help the model better attend to image features, which can help provide more information and improve reasoning.

\section{Methodology} \label{sec:methodology}
In this section, we propose \textbf{Vision-Caption aware Supervised Fine-Tuning (VCASFT)}, a novel supervised training strategy for adapting Vision Language Models (VLMs) towards scientific question-answering tasks. We first introduce supervised fine-tuning of VLMs on scientific multimodal Q\&A datasets and then introduce VCASFT, a modified version of general instruction-tuning.

\subsection{Supervised Fine-Tuning} \label{sft}
Classical supervised fine-tuning enhances pre-trained LLMs on Q\&A datasets through weight optimization within a supervised framework. For a given question $Q$, the model predicts an answer $A_{gen}$, where for each generated answer token $A^{i}_{gen} \in A_{gen}$, cross-entropy loss is computed against the ground truth $A_{real}$. In training, multi-headed attention components $Q, K$, and $V$ are purely textual.

For proficiency in multimodal datasets, VLMs incorporate an image $I$, using a pre-trained encoder $f_{enc}$ to extract features:
\begin{equation}
    X_{I} = f_{enc}(I)
\end{equation}
This modifies the input for multimodal contexts, blending text and image data in the $Q, K, V$ format for cross-attention, enhancing answer generation $A_{gen}$. The model fine-tunes both text and image paths, optimizing for accuracy in relation to the true answer $A_{real}$. Training and zero-shot inference can be represented respectively as:
\begin{equation}
    \text{Train}: Q_t + I_t \longrightarrow A_t
\end{equation}
\begin{equation}
    \text{Zero-Shot Inference}: Q_i + I_i \longrightarrow A_i
\end{equation}

\subsection{Vision-Caption aware Supervised Fine-Tuning (VCASFT)} \label{VCASFT}

Mathverse \citep{zhang2024mathverse} highlighted the limitations of VLMs in extracting mathematical reasoning from images, relying instead on textual cues. To enhance image comprehension, we introduce VCASFT, utilizing an image-captioning model to generate a caption $C$ from image $I$:
\begin{equation}
    z = f_{caption-encoder}(I)
\end{equation}
\begin{equation}
    C = g_{caption-decoder}(z)
\end{equation}
Using Gemini-Pro, captions for images were generated and manually refined to remove inaccuracies and add missing information, ensuring the captions accurately represent the visual content\footnote{Prompts for this task are shared in the Appendix section}. These English captions were translated into Hindi using Gemini Pro Large Language Model and further refined by annotators. The corrected Hindi captions are then used in the VCASFT framework alongside the question text. 

The training process makes the model's attention-module \textbf{better attend} to image features through cues from the textual captions. This helps in enhancing visual comprehension and eventual performance on downstream scientific Q\&A tasks. The modified instruction-prompt used during training is given as follows:
\begin{equation}
    Q^* = \text{Caption:} \sum_{c \in C} c + \text{Question: } Q  
\end{equation}
The enhanced prompt $Q^*$, combining both caption and question, guides the VLM during training and inference:
\begin{equation}
    \text{Train}: Q_t + I_t \longrightarrow A_t
\end{equation}
\begin{equation}
    \text{Zero-Shot Inference}: Q_i + I_i \longrightarrow A_i
\end{equation}

Our approach addresses the fundamental challenge where VLMs struggle to understand complex mathematical diagrams, equations, and geometric figures by providing detailed textual descriptions that serve as interpretive bridges as shown in Figure \ref{fig:VCASFT}. The caption generation process employs specialized prompting strategies designed to capture mathematical symbols, spatial relationships, and numerical contexts that are often missed by standard visual encoders.

When the input question is in English, the VCASFT framework follows a streamlined processing pathway that leverages the native language capabilities of the pre-trained captioning model. The system directly generates detailed English captions from the input image using Gemini-Pro, which ensures mathematical accuracy and completeness. The enhanced prompt is constructed by concatenating the refined English caption with the English question. This unified processing eliminates potential semantic loss, allowing the VLM to maintain consistency between visual descriptions and textual queries throughout the entire reasoning pipeline.

For Hindi language inputs, the framework implements a more complex cross-lingual processing strategy that preserves mathematical semantics while ensuring linguistic accessibility. Initially, detailed English captions are generated from the input image using the same Gemini-Pro model to maintain mathematical precision. The English captions are then translated into Hindi using a specialized translation model fine-tuned on mathematical and scientific terminology. The final enhanced prompt combines the refined Hindi caption with the original Hindi question, enabling native language comprehension while maintaining the mathematical rigor established in the English captioning phase.

\section{Evaluation} \label{experiments}

Based on the methodology described above, we describe our experimental setup. We test our training strategy on the English-language ScienceQA dataset and our proposed Hindi-language dataset HiSciVQA.

\subsection{Dataset} \label{sec:dataset}
In this section we discuss about existing ScienceQA dataset and our HiSciVQA dataset. We provide details about distribution of questions of various types in the dataset, data curation and collection process. We further discuss data augmentation to expand the dataset, followed by details on dataset annotation.

\subsubsection{ScienceQA}
ScienceQA \citep{lu2022learn} serves as a vital benchmark for assessing question-answering models in the scientific domain, featuring a comprehensive array of science-related questions across biology, chemistry, physics, and earth sciences. With its extensive coverage, ScienceQA has become a key resource in scientific Visual Question Answering (VQA). The dataset comprises high-quality multiple-choice questions, including a \textbf{training set of 6,218 samples} and a \textbf{test set of 2,017 samples}, spanning grades 1 to 12 across various subjects. This diversity makes ScienceQA an ideal dataset for benchmarking the effectiveness of our VCASFT technique.

\subsubsection{HiSciVQA}
Identifying a major gap in the availability of question-answer datasets for low-resource languages such as Hindi and advancing the multi-modal language modeling paradigm for scientific Q\&A, this study proposes the creation of a novel multi-modal Hindi Science Visual Question Answering dataset called \textbf{HiSciVQA}. The data curation, annotation, and augmentation process is described briefly with more details in the supplementary.

\subsubsection{Dataset Description}
\begin{figure}[b]
  \includegraphics[width = 0.5\textwidth]{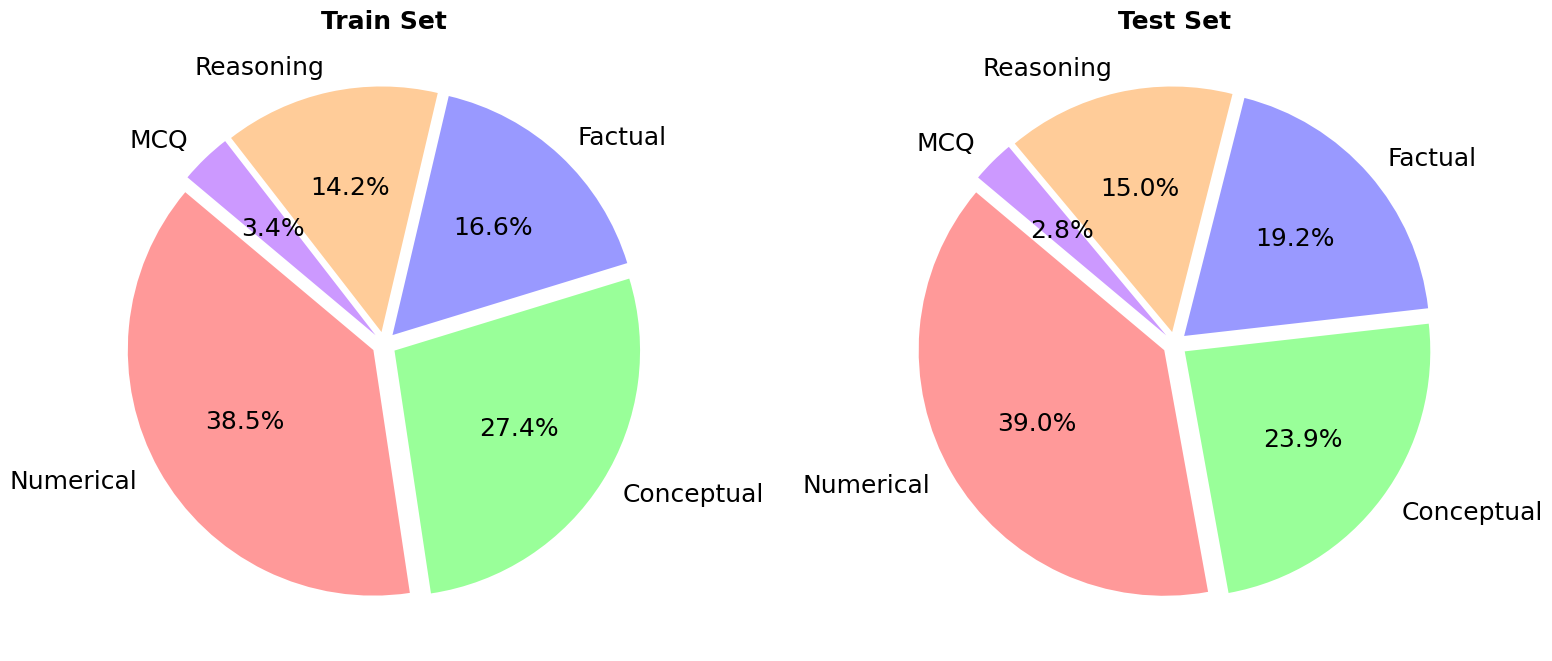}
  \caption{Distribution of train and test splits for HiSciVQA.}
  \label{fig:dist}
\end{figure}

HiSciVQA consists of 2,245 high-quality Hindi-language multi-modal question and answer pairs based on high school science. For each sample in the dataset, there is a question, answer and an accompanying image. The questions in the dataset are divided into five distinct categories:

\begin{enumerate}
    \item \textbf{Numerical Reasoning Based Q\&A:} Questions necessitating algebraic manipulations and mathematical reasoning, where interpreting constant values and measurements from the associated image is crucial.
    \item \textbf{Theoretical Reasoning Based Q\&A:} Questions requiring the comprehension and application of science concepts and theories, without the need for mathematical calculations but relying on conceptual context from images or the model's knowledge.
    \item \textbf{Conceptual Q\&A:} Conceptual application questions focus more on understanding concepts and applying them to a scenario rather than reasoning.
    \item \textbf{Factual Q\&A:} Straightforward questions focusing on knowledge of specific facts or concepts, generally expecting concise, unambiguous answers. Whether the model knows this or not, it cannot be inferred from the image or the text.
    \item \textbf{Multiple Choice Questions (MCQs):} Questions with four options, including one correct answer, supplemented by annotated reasoning and explanation steps.
\end{enumerate}

The train and test's set composition of these five types of questions is given in Figure \ref{fig:dist}. To our knowledge, this is the \textbf{first dataset} in the Hindi language for scientific VQA\footnote{We provide samples for each type of question in the dataset in Appendix}.

\subsubsection{\textbf{Data Curation and Collection}}
Question-answer pairs were collected by scraping government websites, past exams, and other public domain sources, with GPT-4 extracting the relevant questions and their components manually converted to LaTeX. A total of 580 unique multi-modal pairs were curated and divided in a 65-35 split into training and testing sets, containing \textbf{367 and 213} pairs respectively. The training set underwent augmentations to enhance its size and diversity while the test set was reserved.

\begin{figure}
  \includegraphics[width=0.4\textwidth]{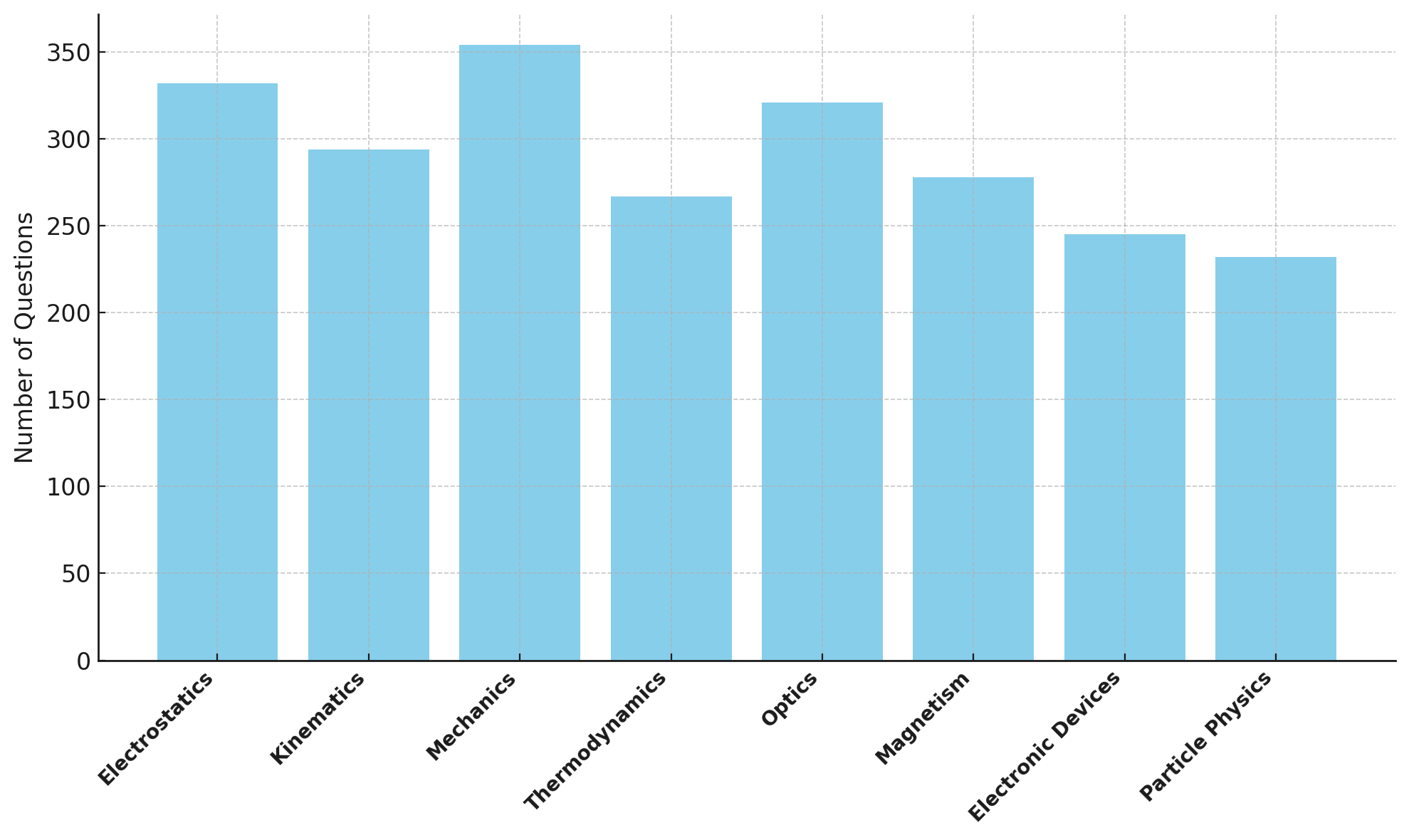}
  \caption{Distribution of number of questions per topic in HiSciVQA dataset.}
  \label{fig:topic_distribution}
\end{figure}

\subsubsection{\textbf{Data Augmentation}}
\vspace{-0.5em}

Traditional data augmentation techniques often misrepresent the context of scientific Q\&A datasets, leading to incorrect reasoning processes. This study introduces a novel data augmentation framework specifically tailored for scientific Q\&A texts, ensuring the integrity of the data is maintained. The steps employed by us as part of the framework are described as follows:

\textbf{Constant Replacement (CR):} This involves providing GPT-4 with a custom prompt to replace all constants in a question with new random values. Additionally, the solution must be updated according to these new values. An annotator supervises the process and rectifies any errors made by GPT-4 to maintain the dataset's high quality.

\textbf{Paraphrasing (Pa):} GPT-4 \citep{achiam2023gpt} and Gemini-Pro \citep{team2023gemini} are employed to rephrase the question while maintaining its original meaning. The prompt used for this task is \emph{"Rephrase the provided question while preserving its symbolic meaning"}. This process alters the wording of the question while keeping the solution unchanged.

Our framework can generate up to $N=10$ candidate questions per original query. A manual selection led to 10 augmented questions for numerical queries and 5-6 for non-numerical ones, contributing to HiSciVQA. This expanded the training set to 2043 entries, detailed in Figure \ref{fig:topic_distribution}. Applicable across languages in the Q\&A domain, this method ensures diverse samples as seen by cosine similarity scores between original and augmented question SBERT embeddings averaging at \textbf{0.45}.

\subsubsection{\textbf{Data Annotation:}}

The first step of the annotation process involved employing LaTeX for text formatting. The annotation team comprised of six native Hindi-speaking engineering undergraduates who excelled in high school science. Measurements and units were added in Hindi for clarity while MathpixOCR, combined with manual methods helped create the image corpus. Each annotator handled about 100 samples, and after augmentation, redistributed them to ensure diversity in the final training set. Lastly, in the augmented samples, reasoning steps and the final answers were checked to ensure correctness with the constants and phrasing of the new question. Incorrect steps were removed and answers were changed accordingly. Independent cross-validators confirmed these final annotations, achieving a \textbf{\emph{Cohen's Kappa score of 0.74}} for the process.

\begin{table}[htbp]
\centering
\caption{Fine-Tuning Parameters.}
\begin{tabular}{@{}ll@{}}
\toprule
\textbf{Parameter}    & \textbf{Description}        \\ \midrule
Method                & Low-Rank Adaptation (LoRA)  \\ 
GPU                   & NVIDIA A100                 \\ 
Optimizer             & Adam     \\ 
LoRA Rank             & 64                          \\ 
LoRA Alpha             & 128                          \\ 
Batch Size            & 8                           \\ 
Epochs                & 3                         \\ 
Learning Rate         & $2 \times 10^{-5}$          \\ \bottomrule
\end{tabular}
\label{tab:finetuning_parameters}
\end{table}

\begin{table*}[]
\centering
\caption{Result Comparison between SFT and VCASFT on ScienceQA dataset.}
\begin{tabular}{cccccccccc}
\toprule
\textbf{Model} & \textbf{Method} & \textbf{Full Test Set} & \multicolumn{3}{c}{\textbf{Subjectwise Eval}} & \multicolumn{3}{c}{\textbf{Gradewise Eval}} \\ & & & \textbf{Social Sci} & \textbf{Natural Sci} & \textbf{Lang. Sci} & \textbf{Lower} & \textbf{Secondary} & \textbf{Higher} \\
\midrule
\multirow{2}{*} {Llava-1.5 7b llama2} & SFT & 82.70 & 89.40 & 78.66 & 77.27 & 86.53 & 77.52 & 80.00 \\
& {VCASFT} & \textbf{88.55} & \textbf{92.93} & \textbf{85.77} & \textbf{88.64} & \textbf{92.4} & \textbf{83.26} & \textbf{100} \\
\midrule
\multirow{2}{*} {Llava-1.6 7b vicuna} & SFT & 84.48 & 90.94 & 80.98 & 70.45 & 87.22 & 80.80 & 80.00 \\
& {VCASFT} & \textbf{88.05} & \textbf{92.93} & \textbf{84.95} & \textbf{88.64} & \textbf{91.88} & \textbf{82.79} & \textbf{100} \\
\midrule
\multirow{2}{*} {Llava-1.6 13b vicuna} & SFT & 86.12 & 92.41 & 82.05 & 88.64 & 88.77 & 82.44 & 100 \\
& {VCASFT} & \textbf{89.19} & {92.28} & \textbf{87.34} & {86.36} & \textbf{91.88} & \textbf{85.60} & {80} \\
\midrule
\multirow{2}{*} {Intern-VL2 Chat Phi-3 3.8B} & SFT & 94.05 & 95.68 & 93.30 & 86.36 & 94.73 & 93.09 & 100 \\
& {VCASFT} & {92.76} & \textbf{97.51} & {89.66} & \textbf{95.45} & \textbf{96.37} & {87.82} & {100} \\
\midrule
\multirow{2}{*} {Intern-VL2 Chat 8B} & SFT &  95.39 & 96.86 & \textbf{94.71} & 88.64 & 95.94 & \textbf{94.61} & 100 \\ & VCASFT & 94 & \textbf{98.04} & 91.56& \textbf{90.91}& \textbf{96.80} & 90& 100 \\
\midrule

\end{tabular}
\label{tab:ScienceQA_Results}
\end{table*}

\subsection{Models and Baseline Experiment Setup} \label{baseline_expt}
\paragraph{\textbf{Benchmarked Models: }} 
We utilized state-of-the-art open-source multimodal language models to address our multimodal scientific Visual Question Answering (VQA) task for the English language. Specifically, we selected the LLaVA family of models, including \textbf{LLaVA-1.5 7B}, \textbf{LLaVA-1.6 7B Vicuna}, and \textbf{LLaVA-1.6 13B Vicuna}. Additionally, we chose two models from the InternVL family: \textbf{InternVL2-Chat 8B} and \textbf{InternVL2-Chat Phi-3 4B}. These models were selected due to their advanced capabilities in vision-language generation tasks, making them well-suited for testing our strategy.

For the Hindi language dataset, we benchmark the state-of-the-art \textbf{PALO series} \citep{maaz2024palo} VLMs, the first multilingual multimodal models which support Hindi. We benchmark all PALO models, from \textbf{MobilePALO} (1.3 billion parameters) to \textbf{PALO-7b and PALO-13b} (7b and 13b parameters respectively), marking their first evaluation on a multilingual, multimodal dataset and demonstrating their linguistic versatility.

\paragraph{\textbf{Baseline Experimental Setup:}} In our baseline experiments, we evaluate models across English and Hindi datasets using Parameter-Efficient Fine-Tuning (PEFT), particularly through Low-Rank Adaptation (LoRA), where the weight matrix $W$ is decomposed into a lower-ranked matrix of rank $r$ to enhance update efficiency during fine-tuning. Both the PALO series and the LLaVA family of models share parameters in this approach, detailed in Table \ref{tab:finetuning_parameters}. The InternVL series also adheres to similar parameters, but differs in that it requires only one epoch of training to converge, compared to multiple epochs for the other models. This set of experiments uses the classic supervised fine-tuning (SFT) as a baseline to compare our proposed training strategy against.

\subsection{VCASFT Experimental Setup} \label{VCASFT_setup}
This section details our experimental framework depicted in Figure~\ref{fig:VCASFT} for the VCASFT strategy. The process begins with generating image captions using the Gemini-Pro Vision model \citep{team2023gemini} in a zero-shot setting tailored for scientific Q\&A applications using caption prompts\footnote{Prompts for generating captions using Gemini are included in the Appendix}. Although Gemini-Pro Vision effectively produces high-quality English captions, its performance in generating Hindi captions is suboptimal, capturing less relevant information. To address this, English captions are translated into Hindi and combined with the original question text into a unified prompt used for model fine-tuning along with the image. The Supervised Fine-Tuning (SFT) approach follows the baseline setup to maintain consistency across experiments.

\begin{table*}[h]
  \centering
  \caption{Text generation evaluation metrics values on the test set of HiSciVQA.}
  \small
  \label{tab:combined_palo_scores}
  \begin{tabular}{ccccccc}
    \toprule
    \textbf{Model} & \textbf{Task} & \textbf{Input Components} & \textbf{Rouge1} & \textbf{Rouge2} & \textbf{RougeL} & \textbf{METEOR} \\
    \midrule
    \multirow{2}{*}{Palo 7b} & SFT & Text, Image & \textbf{0.199} & 0.05 & \textbf{0.168} & \textbf{0.18} \\
                              & VCASFT & Text, Image, Caption & 0.192 & \textbf{0.051} & 0.161 & 0.169 \\
    \midrule
    \multirow{2}{*}{Palo 13b} & SFT & Text, Image  & 0.184 & 0.051 & 0.154 & \textbf{0.191} \\
                               & VCASFT & Text, Image, Caption  & \textbf{0.188} & \textbf{0.051} & \textbf{0.157} & 0.173 \\
    \midrule
    \multirow{2}{*}{Mobile Palo 1.7b} & SFT & Text, Image & \textbf{0.149} & 0.03 & \textbf{0.124} & \textbf{0.15} \\
                                        & VCASFT & Text, Image, Caption  & 0.144 & \textbf{0.037} & 0.12 & 0.143 \\
    \bottomrule
  \end{tabular}
\end{table*}

\subsection{Evaluation Metrics} \label{eval_schema}
\subsubsection{ScienceQA: }ScienceQA consists of multiple-choice questions, each with a single correct answer. Consequently, we utilize a straightforward Accuracy metric for model evaluation on this dataset, where Accuracy is defined as the ratio of the number of correct responses to the total number of questions.

\begin{table*}[ht]
  \centering
  \caption{Scores on Various Question Types of HiSciVQA using SFT and VCASFT.}
  \label{tab:table4}
  \small
  \begin{tabular}{ccccccc}
  \toprule
  Model & Task & Numerical Reasoning & Theoretical Reasoning & Fact-Based & MCQ & Conceptual \\ 
  \midrule
    \multirow{2}{*}{Palo-7b} & SFT & 0.160 & 0.312 & 0.532 & 0.185 & 0.336 \\
    & VCASFT & \textbf{0.188} & \textbf{0.373} & 0.528 & \textbf{0.353} & \textbf{0.532} \\
    \midrule
     \multirow{2}{*}{Palo-13b} & SFT & 0.211 & 0.287 & 0.598 & 0.342 & 0.362 \\
    & VCASFT & 0.164 & \textbf{0.304} & \textbf{0.572} & \textbf{0.358} & \textbf{0.428} \\
    \midrule
     \multirow{2}{*}{MobilePalo-1.7b} & SFT & 0.138 & 0.261 & 0.485 & 0.338 & 0.376 \\
    & VCASFT & \textbf{0.358} & \textbf{0.343} & 0.467 & \textbf{0.345} & \textbf{0.489} \\
    \bottomrule
  \end{tabular}
\end{table*}

\subsubsection{HiSciVQA: } The descriptive nature of the HiSciVQA dataset necessitates the use of traditional text metrics like BLEU \citep{papineni2002bleu}, ROUGE \citep{lin2004rouge}, METEOR \citep{banerjee2005meteor}, and BERTScore \citep{zhang2019bertscore} for initial evaluation. Recognizing the limitations of these metrics due to their focus on n-gram overlaps and only semantic similarity, our study introduces three new evaluation metrics specifically designed to better capture the complexities of scientific Q\&A such as the use of scientific concepts and steps performed for reasoning.

For numerical-answer questions, we introduce the \textbf{Final Answer Accuracy ($S_{FAA}$)}, a binary metric evaluating the model's numerical response within a 2-3\% margin of error. To assess reasoning, the \textbf{Intermediate Steps Score ($S_{ISS}$)} checks the accuracy of intermediate steps against ground-truth answers using GPT-4, with exact matches increasing the correct count. Additionally, the \textbf{Conceptual Similarity Score ($S_{CSS}$)}, based on cosine similarities of SBERT embeddings between the model's and ground truth's concept statements, evaluates conceptual alignment in scientific Q\&A\footnote{Prompts for GPT-4 to calculate these metrics are included in Figures \ref{fig:concept_retrieval_prompt}, \ref{fig:fact_checking_prompt}, \ref{fig:mcq_prompt}, \ref{fig:step_extraction_prompt}, \ref{fig:step_eval_prompt} and \ref{fig:final_answers_prompt} in the Appendix}.

Thus, through this study, we propose a novel evaluation schema using the three metrics defined above and cater to the five distinct types of questions on HiSciVQA:
\paragraph{1. Numerical Reasoning Based Q\&A:} These questions require identifying relevant concepts, performing calculations, and deriving numerical answers. Since intermediate steps can be different for deriving the same answer, it is given the lowest weight of 0.15 while the highest of 0.5 is allocated to the accuracy of the final answer.
\begin{equation} \small
        \mathbf{S_{Num} = 0.5\cdot S_{FAA} + 0.15\cdot S_{ISS} + 0.35\cdot S_{CSS}}
\end{equation}

\paragraph{2. Theoretical Reasoning Based Q\&A}: Unlike numerical questions, these involve theoretical reasoning to arrive at an answer without necessitating a numerical final answer. Therefore, we adjust our evaluation to emphasize reasoning and conceptual understanding, excluding $S_{FAA}$ due to the absence of numerical answers. A final weighted score is reported as follows:
    \begin{equation} \small
        \mathbf{S_{Theoretical} = 0.2\cdot S_{ISS} + 0.8\cdot S_{CSS}}
    \end{equation}
Again, as steps can be different while the conceptual grounding needs to be accurate, a ratio of 4:1 is maintained between the concept and Step scores through our weights.
\paragraph{3. Conceptual Q\&A:} These questions, centered on concept application and understanding, are assessed using the \textbf{Conceptual Similarity Score ($S_{CSS}$)} to gauge the model's comprehension of scientific principles.
\paragraph{4. Factual Q\&A:} Questions testing factual accuracy are evaluated with a binary n-gram overlap metric, supplemented by human annotator verification.
\paragraph{5. MCQs (Multiple Choice Questions):} MCQs are approached as a classification task with a simple accuracy metric, determining whether the model selects the correct answer from four options.

\section{Results and Discussion} \label{results}
In this section, we present our scores on the experiments described in the Experiments section. We start by showing values on general text generation metrics and then perform a detailed discussion on each question type in the test set comparing performance based on the novel evaluation schema.

\subsection{Results on ScienceQA}\label{sec:eng_results}
Table \ref{tab:ScienceQA_Results} presents the results of various multi-modal LLMs under SFT and VCASFT settings. The overall accuracy improves from the LLaVA-1.5 7B model to the Intern-VL 8B model as the finetuning progresses on ScienceQA\footnote{Error analysis can be found in the  Section \ref{sec:error_analysis}}.

The Intern-VL2 Chat 8B model achieves the highest score of 95.39 with SFT, excelling across all subjects, especially in Social and Natural Science, and performing well across all grade levels, particularly in the Higher Grade category. The Intern-VL 7B model also performs strongly with a score of 94.00, showing consistent results across subjects and grades.

The VCASFT method enhances performance, particularly for models with weaker Vision Encoders that use MLP connectors between vision and language. For example, Llava-1.5 7B LLaMA2 and Llava-1.6 Vicuna models showed significant gains in tasks like Natural Science and Higher Grade evaluations, where image-text alignment is critical. VCASFT improves their visual-textual understanding through fine-tuned captions.

In contrast, models like InternVL Chat 8B, which have advanced Vision Encoders and LLM-based vision-language connectors, saw limited benefits. These models already extract rich visual features, and VCASFT may introduce redundancy and confusion for models that hinder generalization. Thus, VCASFT is more effective for models with weaker Vision Encoders but less so for those with advanced architectures.

\begin{figure*}[ht]
    \centering
    \begin{subfigure}{0.19\textwidth}
        \centering
        \includegraphics[width=\linewidth]{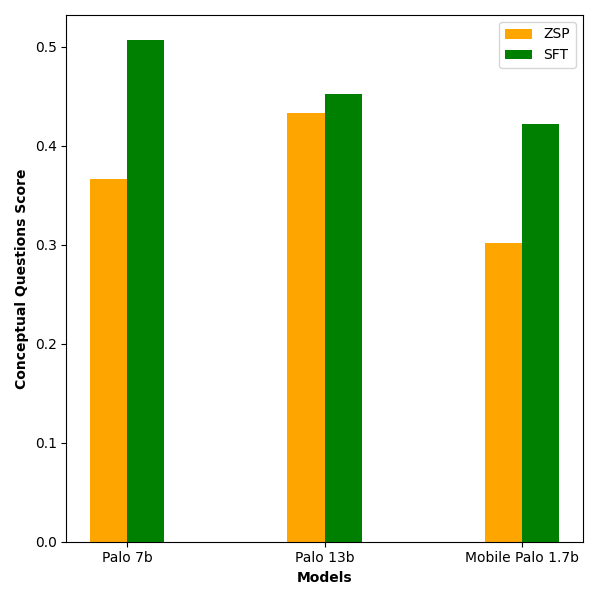}
        \caption{Conceptual Questions}
        \label{fig:conceptual_ablation}
    \end{subfigure}
    \hfill
    \begin{subfigure}{0.19\textwidth}
        \centering
        \includegraphics[width=\linewidth]{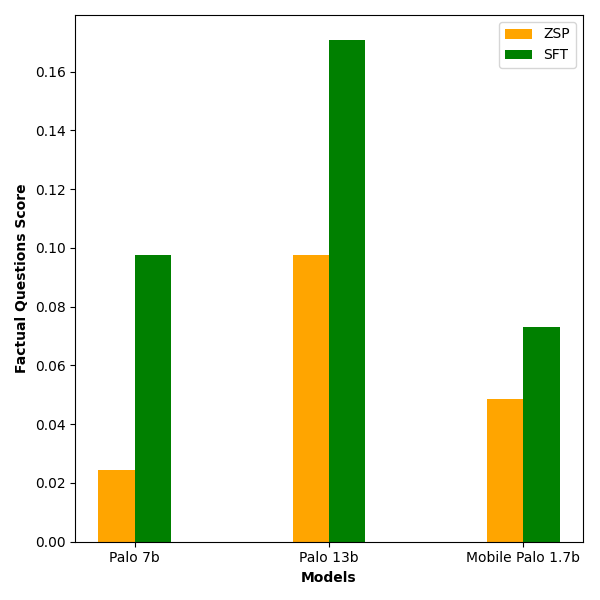}
        \caption{Factual Question}
        \label{fig:fact_ablation}
    \end{subfigure}
    \hfill
    \begin{subfigure}{0.19\textwidth}
        \centering
        \includegraphics[width=\linewidth]{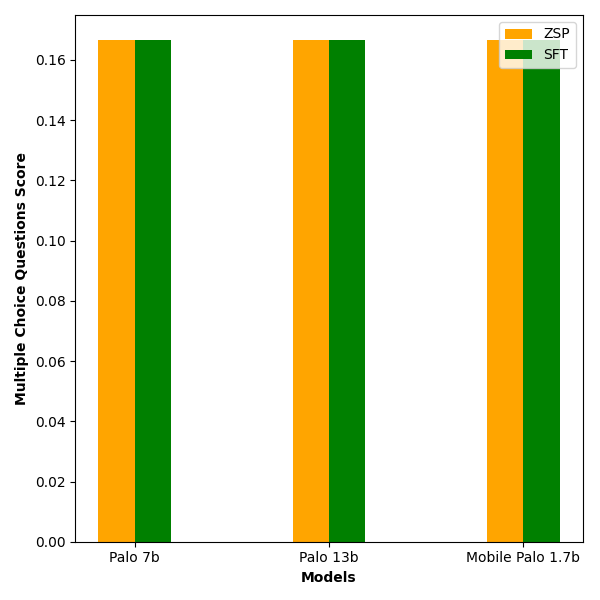}
        \caption{Multi-Choice Questions}
        \label{fig:mcq_ablation}
    \end{subfigure}
    \hfill
    \begin{subfigure}{0.19\textwidth}
        \centering
        \includegraphics[width=\linewidth]{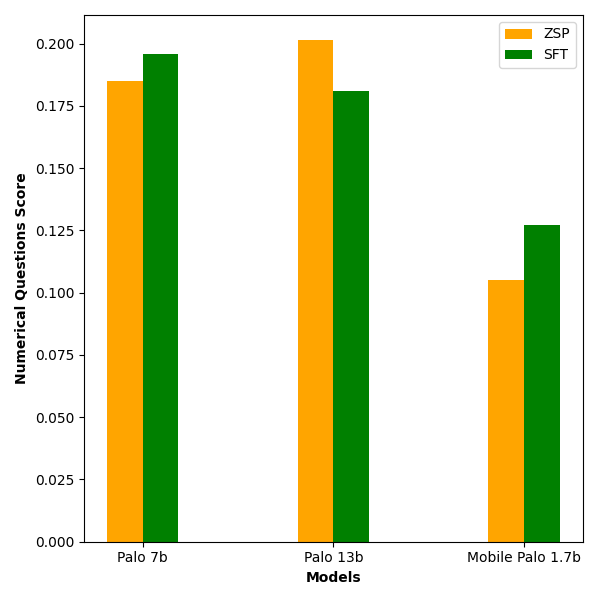}
        \caption{Numerical Questions}
        \label{fig:numerical_ablation}
    \end{subfigure}
    \hfill
    \begin{subfigure}{0.19\textwidth}
        \centering
        \includegraphics[width=\linewidth]{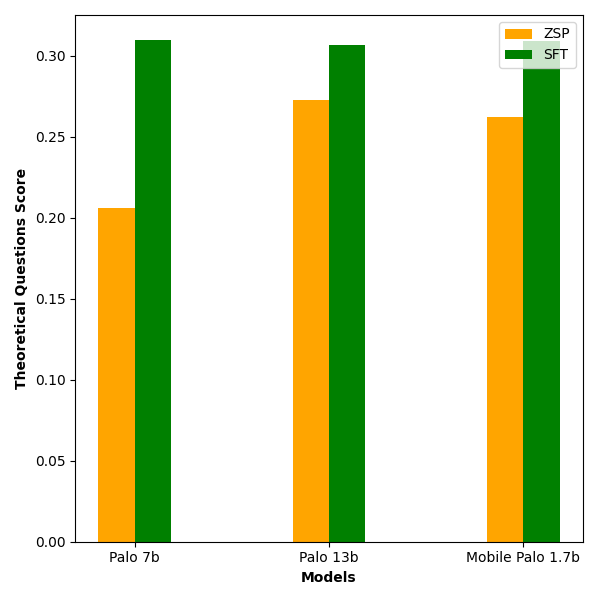}
        \caption{Theoretical Questions}
        \label{fig:theoretical_ablation}
    \end{subfigure}
    \caption{Results on Ablation Study.}
    \label{fig:ablation_results}
\end{figure*}

\subsection{Results on HiSciVQA} \label{sec:hindi_results}
Table \ref{tab:combined_palo_scores} presents a performance analysis of three PALO models on Hindi language Q\&A tasks, using metrics like Rouge and METEOR within the VCASFT framework. The impact of including captions during the fine-tuning process was analyzed with strong trends indicating better performance by VCASFT. In Table \ref{tab:table4} we present scores on various types of questions in HiSciVQA and provide a performance comparison. We subsequently discuss the results obtained using our evaluation schema on HiSciVQA and an ablation study on our methodology in Section \ref{sec:ablation}. Our results on the various question types are detailed as follows:

\paragraph{Conceptual Q\&A:} SBERT scores in Table \ref{tab:table4} demonstrate an improvement in model performance with caption integration, notably with the PALO 7b model achieving the highest score of 0.532. This indicates that captions enhance comprehension and accuracy in scientific Q\&A applications.

\paragraph{Multiple-Choice Questions:} Table \ref{tab:table4} shows that incorporating captions through Supervised Fine-Tuning (SFT) enhances model precision in MCQs by improving the integration of visual and textual data, a benefit observed consistently across various model architectures with increases of around 100\% when looking at the Palo-7b model.
\paragraph{Fact-Based Q\&A:} Table \ref{tab:table4} does not show a significant performance boost which could be attributed to the model's pre-training knowledge base being affected by our training strategy. Improvement on this would require a greater pre-training endeavour.
\paragraph{Theoretical Reasoning Q\&A:} Table \ref{tab:table4} shows that including captions raises the score from 0.312 to 0.373 for Palo-7b, demonstrating how captions enhance model understanding of theoretical constructs and helps produce better steps and reasoning. 
\paragraph{Numerical Reasoning Q\&A:} It shows that including captions improves numerical reasoning scores across models in the VCASFT task, as detailed in Table \ref{tab:table4}. This enhancement underscores that captions boost contextual understanding and model reasoning, allowing for better performance without increasing computational complexity. This is noticeably better for a smaller model like MobilePalo-1.7b, indicating our strategy is effective at augmenting smaller VLMs with weaker vision encoders.

\section{Ablation Study} \label{sec:ablation}

Our ablation study carefully explores the effects of omitting images during fine-tuning. We specifically assess the model's performance when provided only with question text and image captions, aiming to understand the significance of image components in the VCASFT framework. As illustrated in Figure \ref{fig:ablation_results}, fine-tuning the VCASFT framework significantly enhances performance across all question types compared to zero-shot prompting, with a particularly notable improvement of almost 15 points in factual question types across all three model types.
We observe similar gains in conceptual, numerical, and reasoning-based questions, which often rely on information derived from images. Despite the absence of images, models were able to answer questions using only captions, highlighting the critique that Vision Language Models tend to overlook image-based cues. Instead, well-crafted captions can more effectively convey the information needed for downstream tasks.
Our results also reveal a significant performance gap between Supervised Fine-Tuning (SFT) and Zero-Shot Prompting (ZSP), which widens further when no captions are provided, emphasizing the critical role of captions in visual question-answering. These findings highlight the considerable benefits of multimodal integration and underscore the importance of textual caption-based data in optimizing model performance and effectively extracting information from images.

\section{Dataset Details} \label{sec:data}
This section outlines the annotation process and presents samples from the HiSciVQA dataset. Figure \ref{fig:annotaed_sample} illustrates how the datasets were annotated from the source, where source questions and answers were converted into LaTeX format to serve as input for the model during fine-tuning.\footnote{Figures \ref{fig:mcq_sample_image}, \ref{fig:numerical_sample_image}, \ref{fig:reasoning_sample_image}, \ref{fig:factual_sample_image} and \ref{fig:conceptual_sample_image} in the Appendix section showcase samples from our dataset for each of the five types of questions.} We include the question, image, and the corresponding solution from HiSciVQA.

\section{Conclusion}

In this study, we introduced HiSciVQA, a pioneering Hindi dataset focused on high-school-level science Q\&A, aimed at addressing the challenges of Vision Language Models (VLMs) in low-resource languages. We developed VCASFT to enhance VLM performance by using image-generated captions as contextual cues, improving model comprehension and answer generation. Our work significantly advances educational AI and multimodal learning for low-resource languages. We employed a comprehensive evaluation schema beyond traditional accuracy metrics, providing insights into VLMs' understanding and reasoning abilities. This study lays the groundwork for future research to improve multimodal models through innovative learning strategies and highlights the need for diverse language datasets in scientific education.

\section{Limitations}

One limitation of our study is the size of the HiSciVQA dataset, which consists of only 2,245 samples of multimodal visual question answering (VQA) in Hindi. To enhance low-resource multimodal scientific VQA tasks, we need to curate additional data. The VCASFT framework, designed to improve the performance of vision-language models (VLMs) on scientific VQA tasks, has shown promising results with smaller VLMs that possess limited vision encoding capabilities and vision-language connectors. In the future, it will be essential to integrate additional information to generalize this method for broader application across various VLMs in Scientific VQA tasks. This work establishes a foundation for further research in this emerging field.

\section{Ethics Statement}

We provide all the hyperparameters used in our experiments and the prompts used for evaluation for reproducibility. The Vision Language Models employed for finetuning on ScienceQA and HiSciVQA were obtained from publicly available repositories. We recognize the potential for inherent biases in the generated outputs. The HiSciVQA dataset used for training was sourced from NCERT textbooks, which are widely utilized for high school education across India. This ensures the representation of a broad demographic within the country.

\begin{acks}
Rajiv Ratn Shah is partly supported by the Department of Computer Science \& Engineering, Infosys Center for AI, the Center of Design and New Media, and the Center of Excellence in Healthcare at IIIT Delhi.
\end{acks}

\bibliographystyle{ACM-Reference-Format}
\balance
\bibliography{sample-base}

\appendix
\section{Appendix}





\subsection{Prompt Details} \label{sec:prompt_details}
\begin{figure*}[t]
  \includegraphics[width=0.6\textwidth]{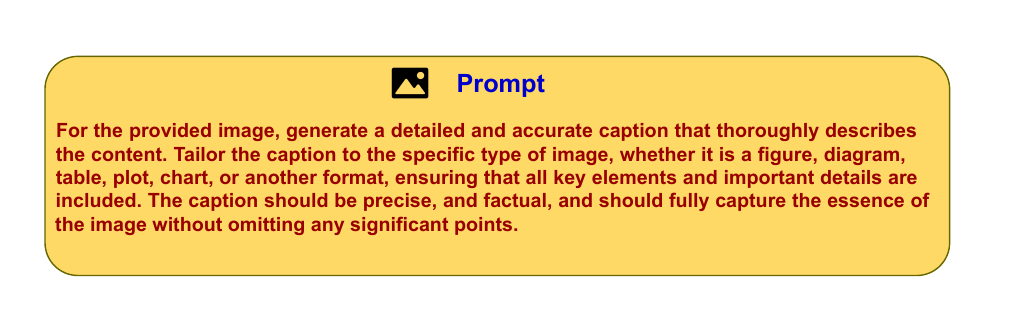}
  \caption{Captioning prompt used to generate detailed image caption.}
  \label{fig:caption_prompt}
\end{figure*}

\begin{figure*}[t]
  \includegraphics[width=0.6\textwidth]{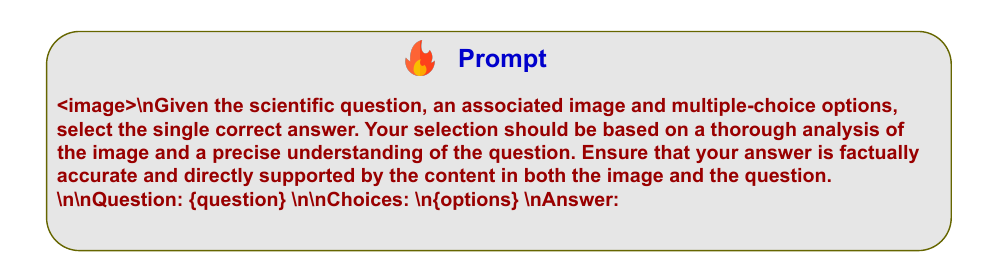}
  \caption{Finetuning prompt.}
  \label{fig:finetuning_prompt}
\end{figure*}

We employ the Gemini Flash 1.5 model to generate precise captions using carefully designed prompts. The image below illustrates the prompt we use to obtain the caption. Figure \ref{fig:caption_prompt} displays the actual prompt provided to the Gemini model for generating captions for an input image. Figure \ref{fig:finetuning_prompt} shows the exact prompt setup used for fine-tuning our model with VCASFT.

\begin{figure*}[]
  \includegraphics[width=0.6\textwidth]{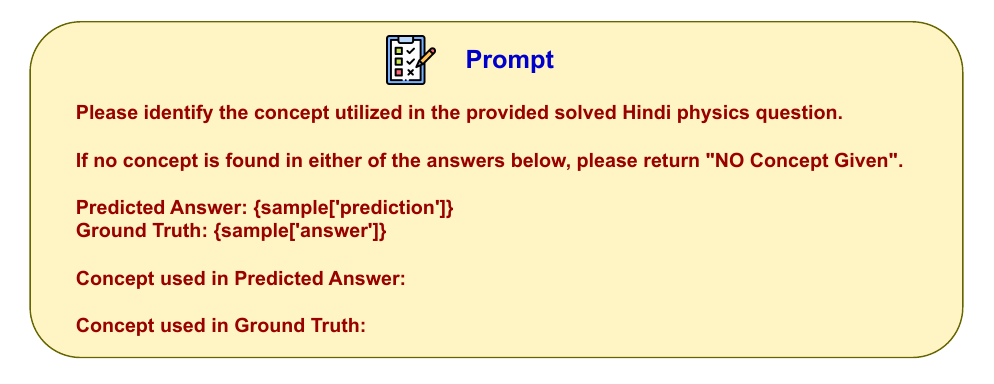}
  \caption{Concept retrieval prompt for predicted and ground truth answer.}
  \label{fig:concept_retrieval_prompt}
\end{figure*}

\begin{figure*}[]
  \includegraphics[width=0.6\textwidth]{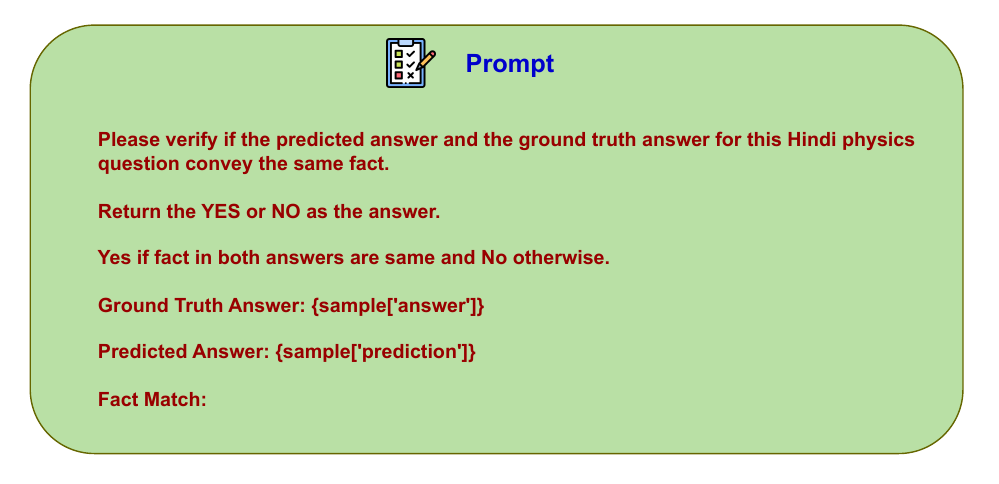}
  \caption{Fact checking prompt for predicted and ground truth answers.}
  \label{fig:fact_checking_prompt}
\end{figure*}

\begin{figure*}[]
  \includegraphics[width=0.6\textwidth]{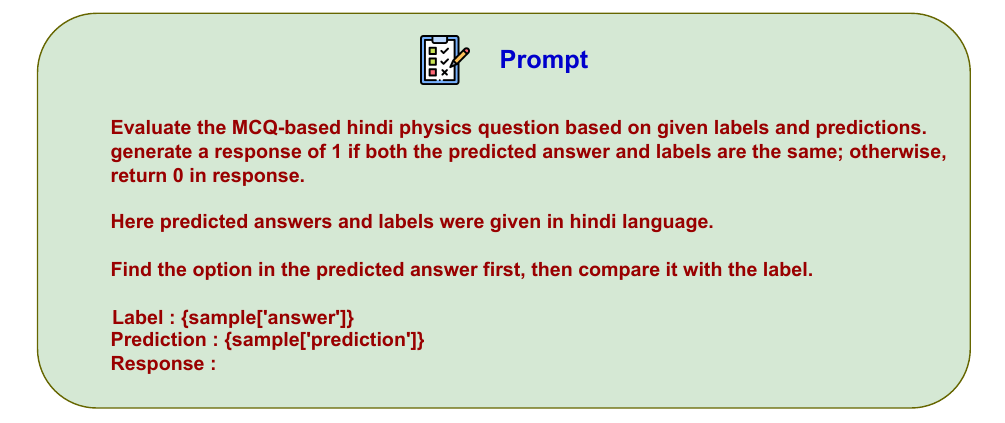}
  \caption{MCQ evaluation prompt.}
  \label{fig:mcq_prompt}
\end{figure*}

\begin{figure*}[]
  \includegraphics[width=0.6\textwidth]{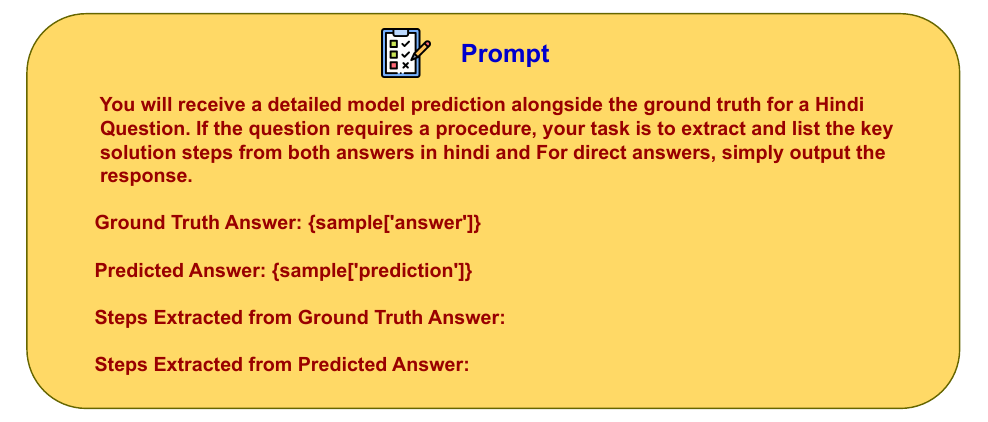}
  \caption{Step extraction prompt from predicted and ground truth answers.}
  \label{fig:step_extraction_prompt}
\end{figure*}

\begin{figure*}[t]
  \includegraphics[width=0.6\textwidth]{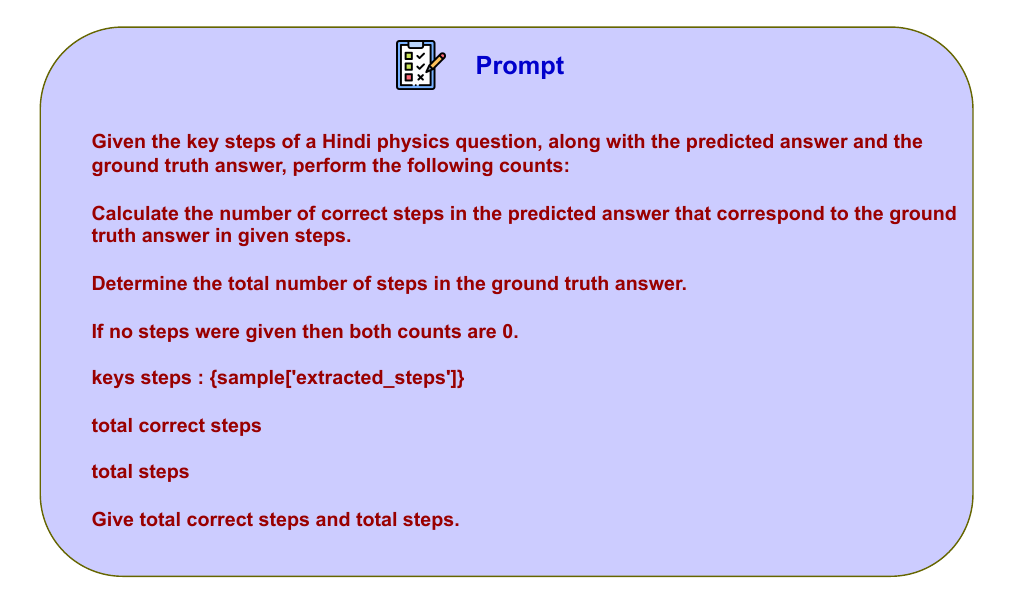}
  \caption{Extracted steps evaluation prompt.}
  \label{fig:step_eval_prompt}
\end{figure*}

\begin{figure*}[t]
  \includegraphics[width=0.6\textwidth]{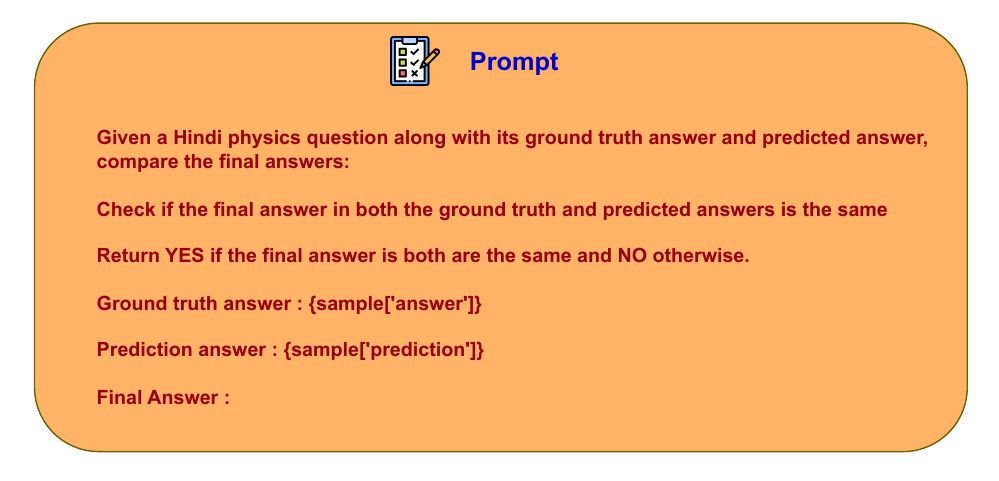}
  \caption{Final answer evaluation Prompt.}
  \label{fig:final_answers_prompt}
\end{figure*}

Prompts for GPT-4 to calculate our evaluation metrics described in Section \ref{eval_schema} are shown in Figures \ref{fig:concept_retrieval_prompt}, \ref{fig:fact_checking_prompt}, \ref{fig:mcq_prompt}, \ref{fig:step_extraction_prompt}, \ref{fig:step_eval_prompt} and \ref{fig:final_answers_prompt}.

\begin{figure*}[htbp]
  \includegraphics[width=0.75\textwidth]{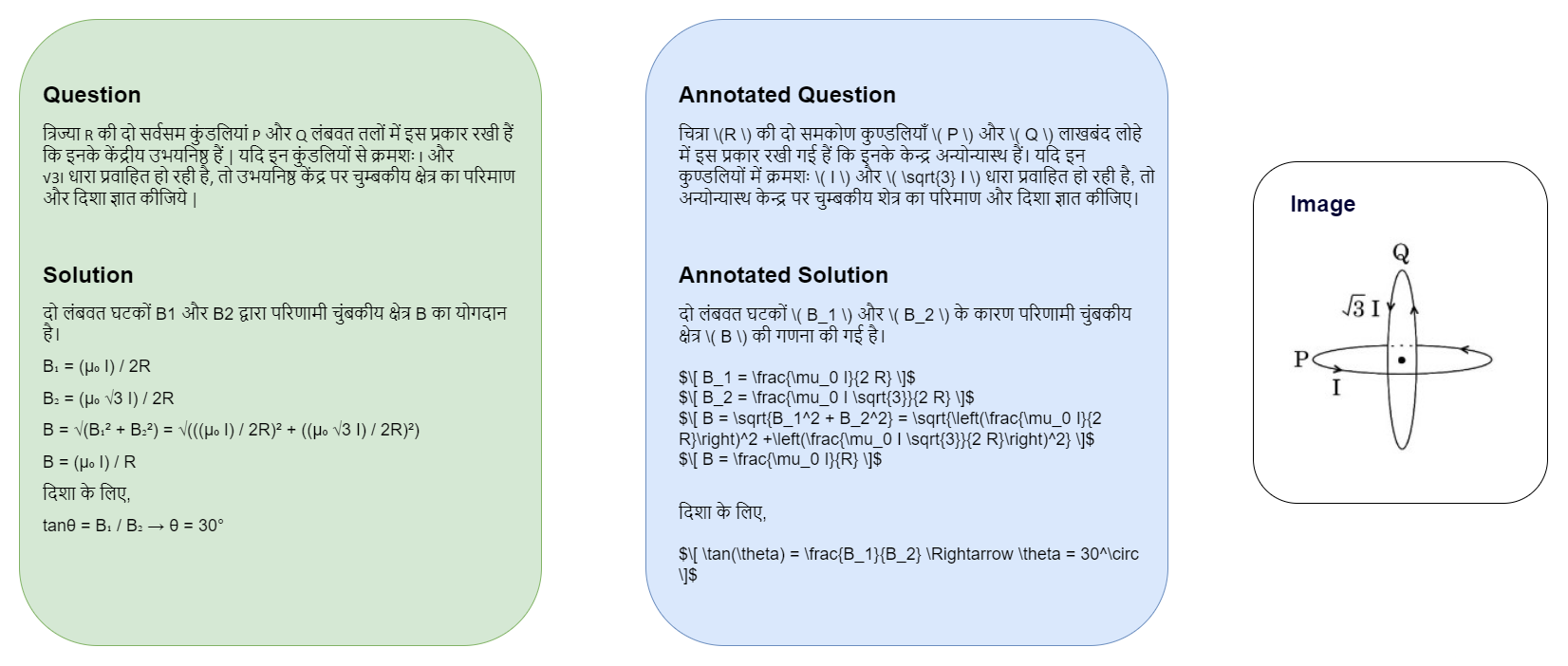}
  \caption{Example of annotated sample from HiSciVQA.}
  \label{fig:annotaed_sample}
\end{figure*}

\begin{table*}[htbp]
  \centering
  \caption{Topics and Subtopics in the HiSciVQA dataset.}
  \label{table:topics}
  \begin{tabular}{p{0.25\linewidth}p{0.65\linewidth}} 
    \toprule
    \textbf{Topic} & \textbf{Subtopics} \\
    \midrule
    Electrostatics \& Current Electricity & Electric Potential, Electric Field, Ohm’s Law, Kirchhoff’s Laws, Combinations of Resistors, Electrical Instruments \\
    Kinematics & Velocity-Time, Acceleration, Rotational Motion, Gravitation, Motion in a Straight Line, Motion in a Plane, Periodic Motion, Wave Motion \\
    Mechanics & Law of Motion, Work, Power, Force, Law of Motion \\
    Thermodynamics & Laws of Thermodynamics, Thermal Equilibrium, Heat Transfer, Temperature, Reversible and Irreversible Processes, Kinetic Theory of Gases \\
    Optics & Ray Optics (Reflection, Refraction, Lenses, and Mirrors) and Wave Optics (Young Double Slit, Huygen's Principle) \\
    Magnetism & Magnetic Field, Hysteresis, Permeability, Electromagnets \\
    Electronic Devices & Semiconductors, Logic Gates, Diode \\
    Particle Physics & Atomic Structure, Nuclei, Isotopes \\
    \bottomrule
  \end{tabular}
\end{table*}

\subsection{Error Analysis} \label{sec:error_analysis}
\begin{figure*}
    \centering
    \includegraphics[width=0.5\textwidth]{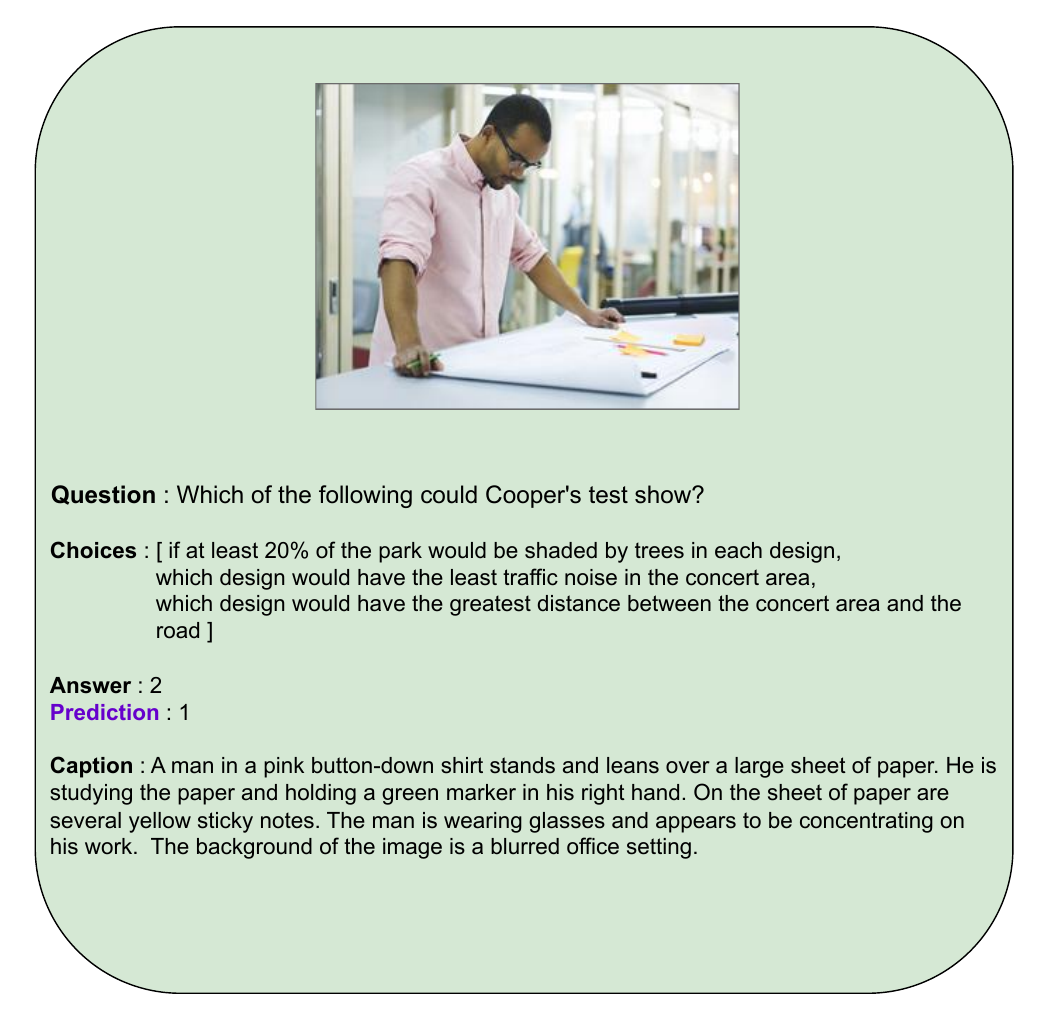}
    \caption{Error example on InternVL 8B model with VCASFT.}
    \label{fig:error_1}
\end{figure*}

\begin{figure*}
    \centering
    \includegraphics[width=0.5\textwidth]{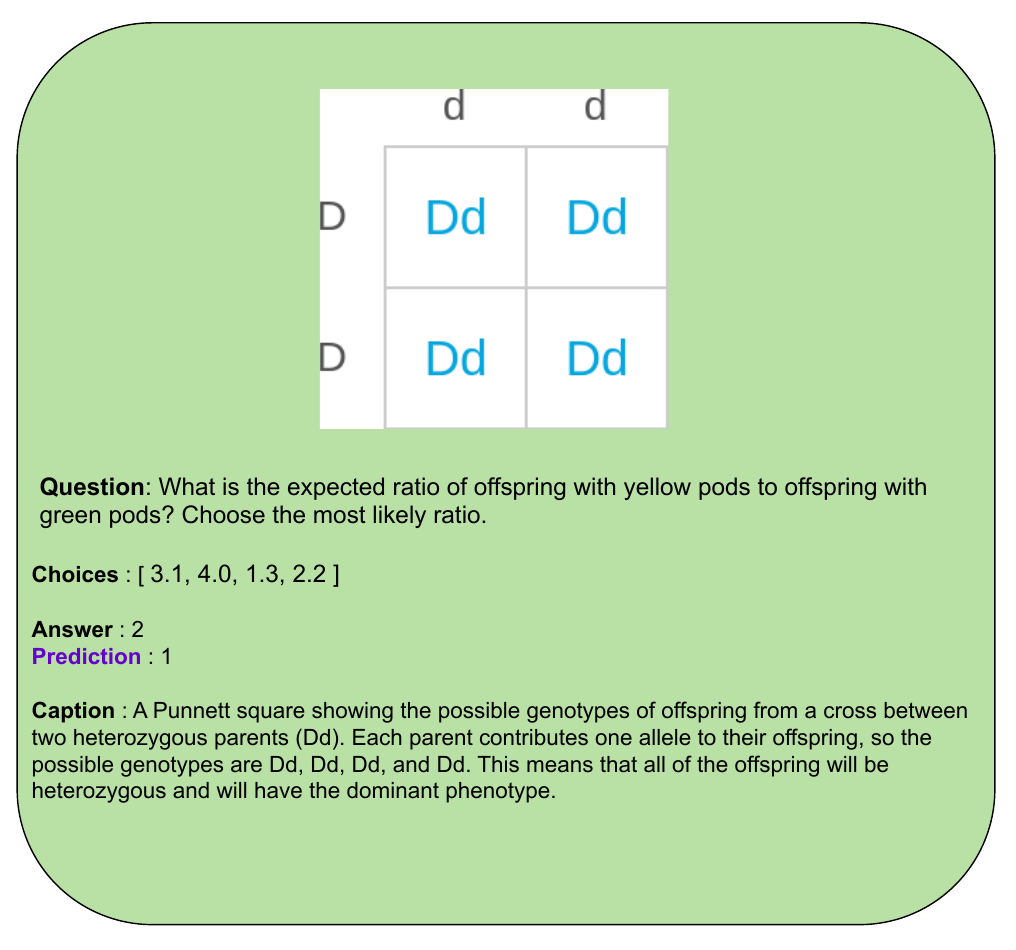}
    \caption{Error example on InternVL Chat 4B Phi3 model with VCASFT.}
    \label{fig:error_2}
\end{figure*}

\begin{figure*}
    \centering
    \includegraphics[width=0.5\textwidth]{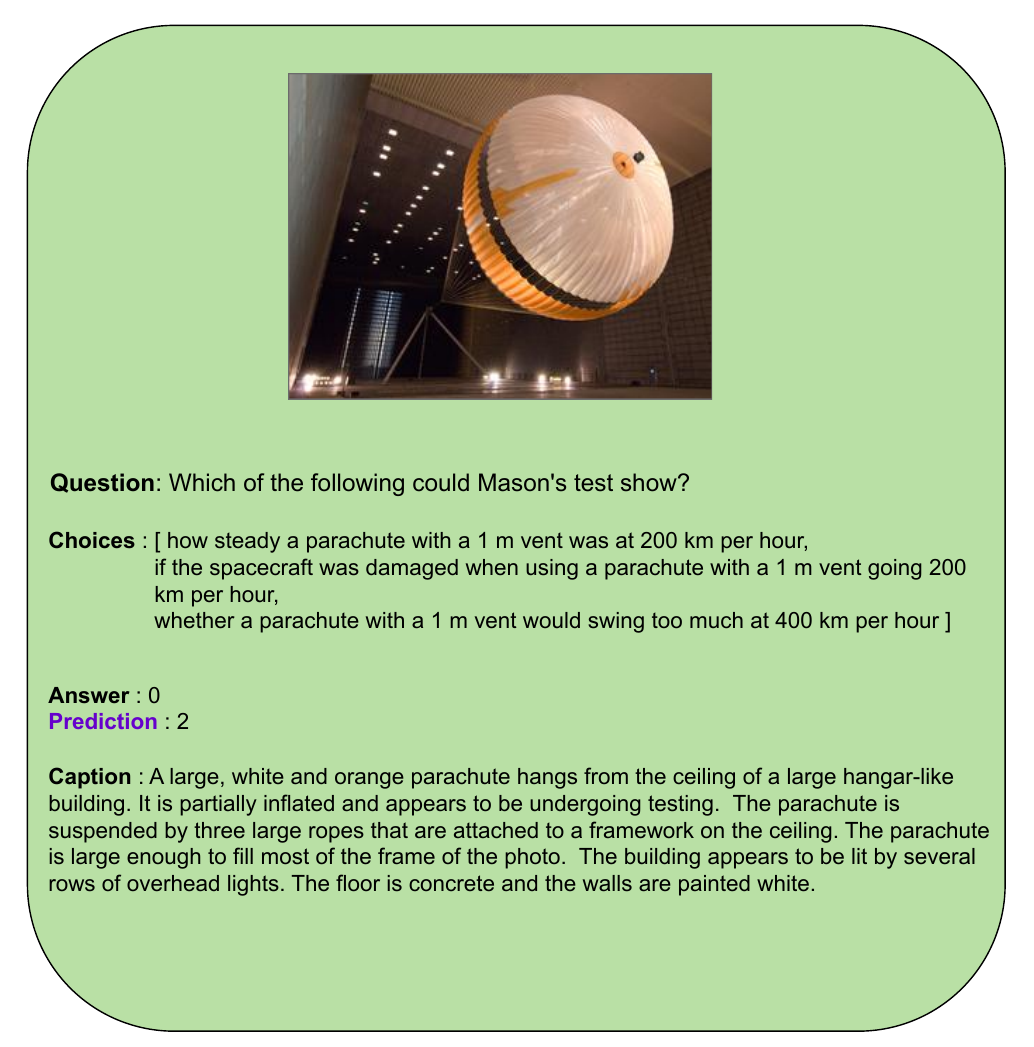}
    \caption{Error example on InternVL Chat 4B Phi3 model with VCASFT.}
    \label{fig:error_3}
\end{figure*}

This section presents the error analysis of the VCASFT experiments on the Intern-VL family of models. As shown in Figure \ref{fig:error_1}, the model made an incorrect prediction because the image did not provide the necessary information to answer the given question accurately. Consequently, the model had to rely on assumptions, and the additional image captions provided in the figure did not assist in arriving at the correct answer due to a misalignment between the image and the question. Similarly, in Figures \ref{fig:error_2} and \ref{fig:error_3}, the images did not directly convey the information needed to correctly answer the questions, leading the models to make errors on such samples from the ScienceQA test set. These inherent issues of misalignment between question and image in the dataset were not adequately addressed by the model, even after caption-aware fine-tuning.

\begin{figure*}[h]
    \centering
    \includegraphics[width=0.6\textwidth]{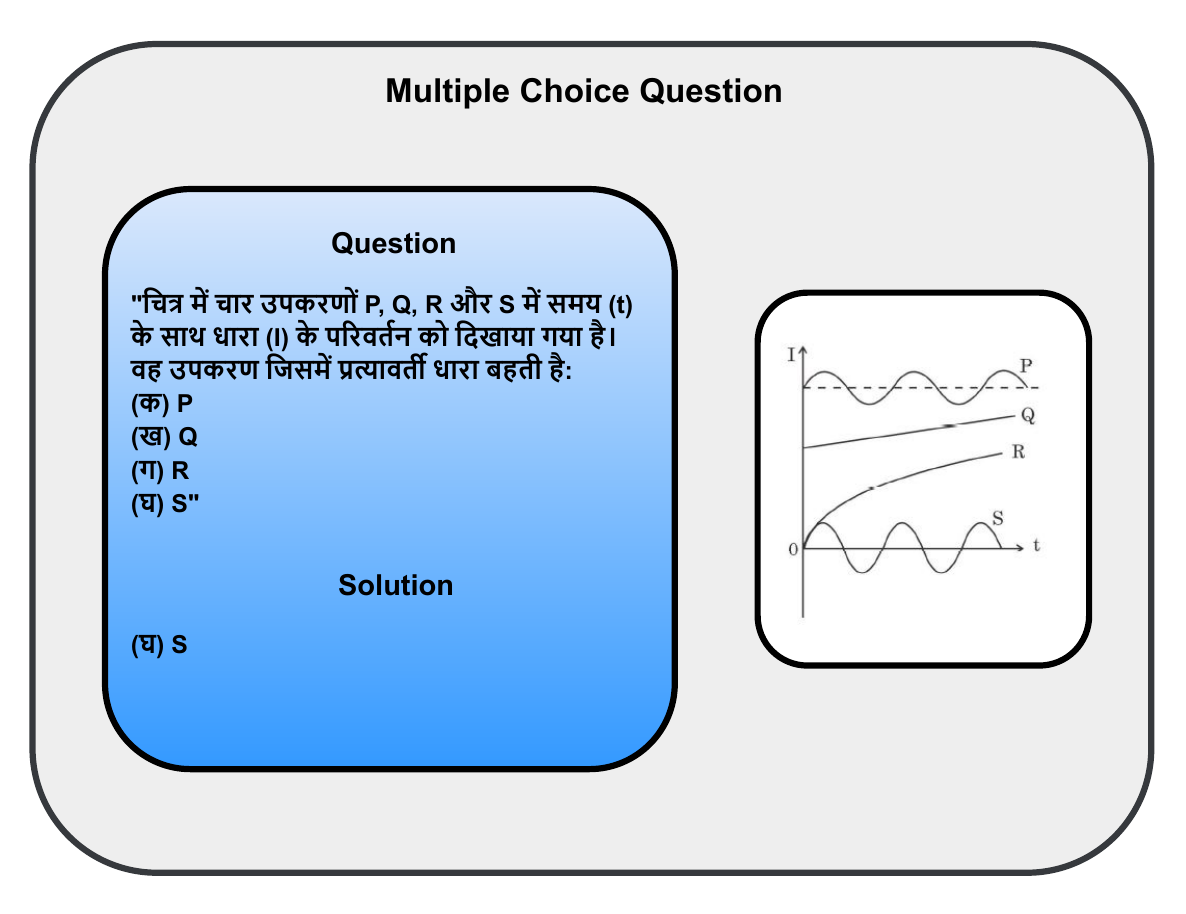}
    \caption{Sample of MCQ type question from HiSciVQA.}
    \label{fig:mcq_sample_image}
\end{figure*}

\begin{figure*}[h]
    \centering
    \includegraphics[width=0.6\textwidth]{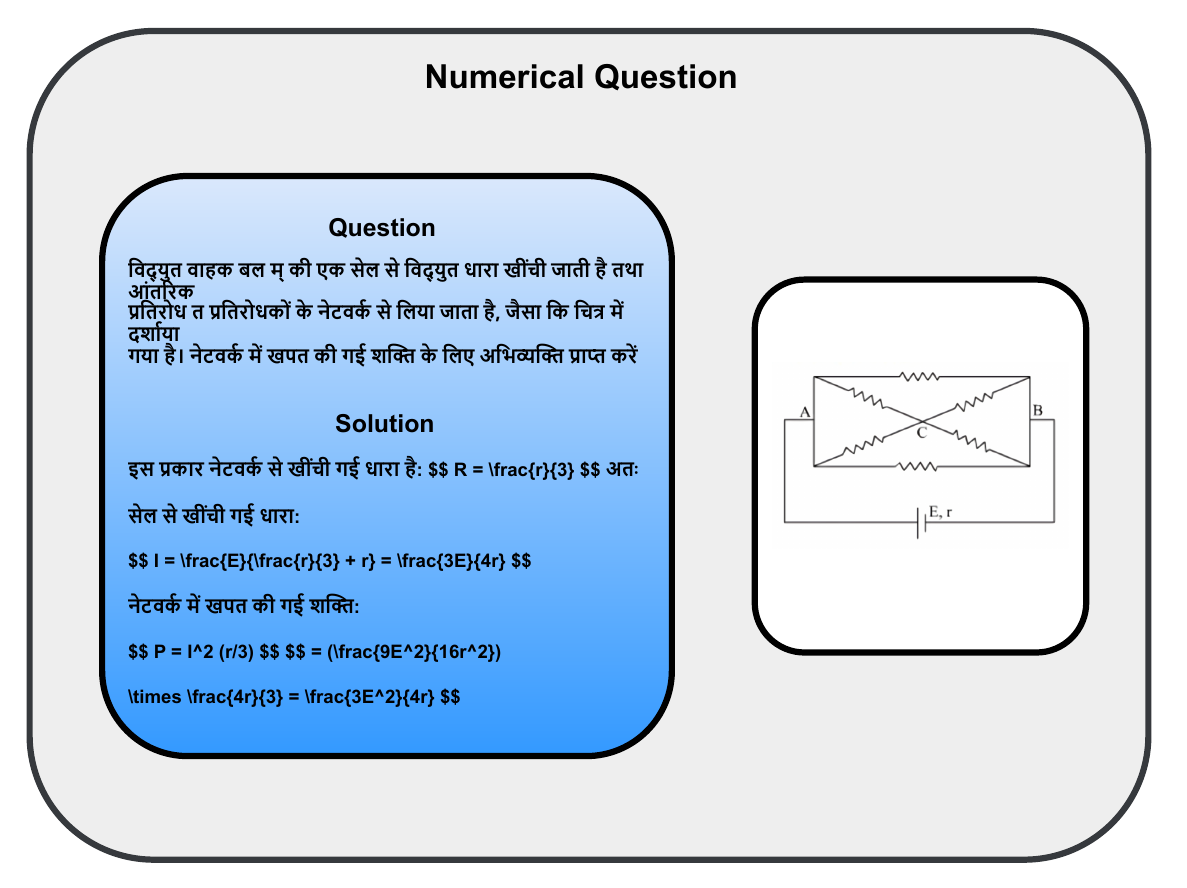}
    \caption{Sample of numerical type question from HiSciVQA.}
    \label{fig:numerical_sample_image}
\end{figure*}

\begin{figure*}[h]
    \centering
    \includegraphics[width=0.6\textwidth]{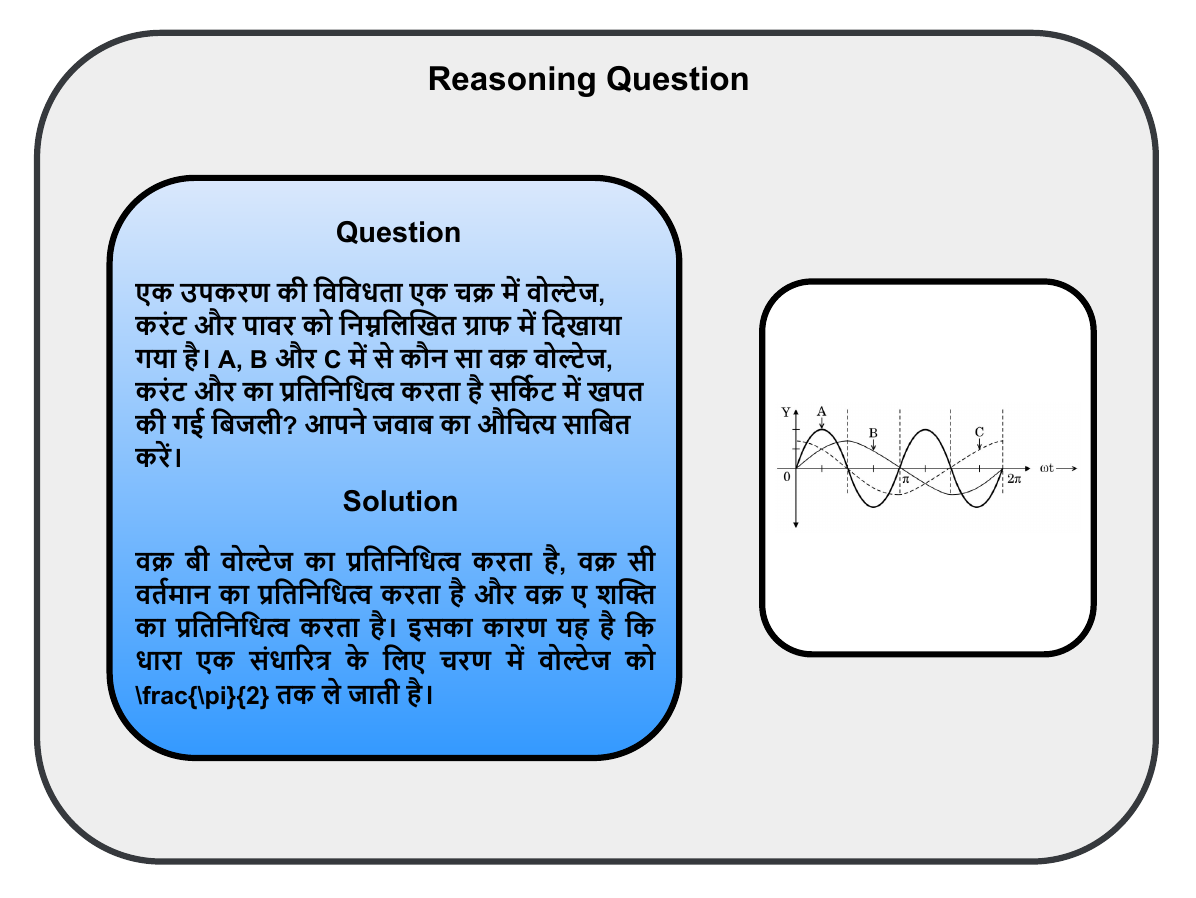}
    \caption{Sample of reasoning type question from HiSciVQA.}
    \label{fig:reasoning_sample_image}
\end{figure*}

\begin{figure*}[h]
    \centering
    \includegraphics[width=0.6\textwidth]{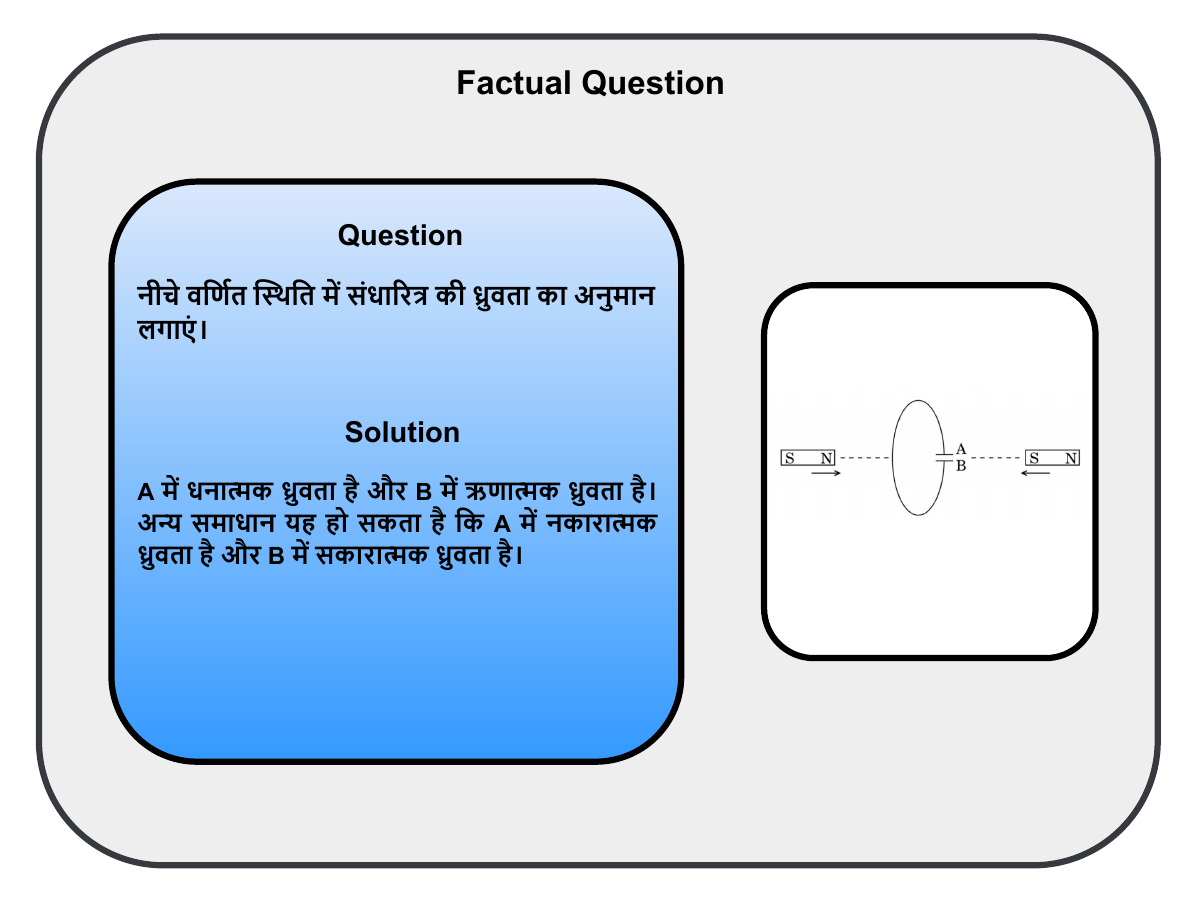}
    \caption{Sample of factual type question from HiSciVQA.}
    \label{fig:factual_sample_image}
\end{figure*}

\begin{figure*}[h]
    \centering
    \includegraphics[width=0.6\textwidth]{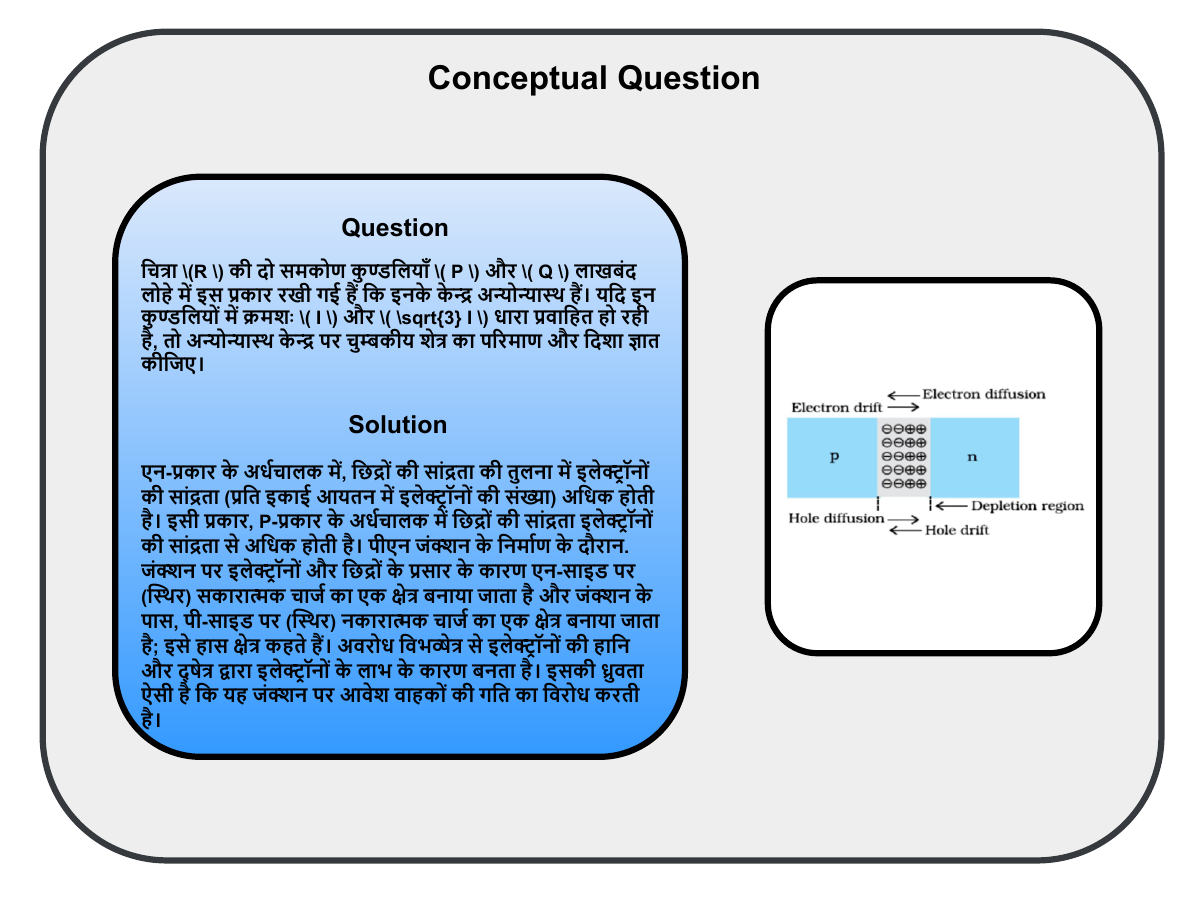}
    \caption{Sample of conceptual type question from HiSciVQA.}
    \label{fig:conceptual_sample_image}
\end{figure*}


\begin{figure*}[]
  \includegraphics[width=0.6\textwidth]{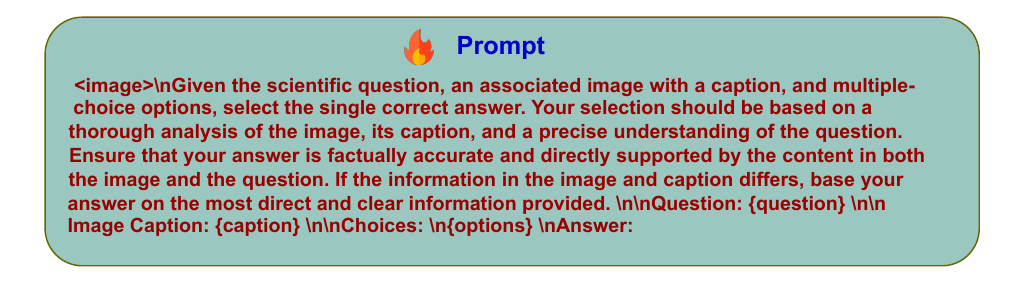}
  \caption{Finetuning with captioning prompt.}
  \label{fig:finetuning_caption_prompt}
\end{figure*}

\end{document}